# Tolerance-Dependent Inspection Disagreement Between a Fixed CMM and a Portable Articulated-Arm CMM

**Md Manjurul Ahsan***, Hamidreza Samadi, Shivakumar Raman

School of Industrial and Systems Engineering, University of Oklahoma,
Norman, OK 73019, USA

Emails: ahsan@ou.edu (M.M. Ahsan); hrs.samadi@gmail.com (H. Samadi); raman@ou.edu (S. Raman)

## Abstract

Fixed coordinate measuring machines (CMMs) and portable articulated-arm CMMs are often assigned to the same inspection task, but their nominal accuracy specifications do not show whether a change of instrument will preserve the disposition of a part. The question is not simply how far the two results differ, but whether that difference crosses the tolerance boundary. We examined this issue with recorded measurements of cylindrical, cubic, and spherical features under nominal 20 °C and 30 °C conditions. Repeated records and two roughness profiles without sufficient acquisition information were removed, leaving six dimensional and four form profiles. For each dimensional feature, the distances of the two system means from nominal define the exact tolerance interval in which the systems receive opposite direct labels. The fixed-CMM stream was approximately 11.2 µm higher than the articulated-arm stream at both conditions. All four form profiles fell on opposite sides of the recorded 10 µm upper limit. The dimensional disagreement intervals also overlapped strongly; their mean widths were 6.573 µm at 20 °C and 4.995 µm at 30 °C. The results clarify why an average difference between instruments is not, by itself, a measure of substitution risk. The proposed tolerance map identifies the feature-tolerance combinations for which instrument choice can change the recorded inspection label and, therefore, where a controlled equivalence study and a task-specific uncertainty budget are needed before substitution.

**Keywords:** coordinate metrology; articulated-arm CMM; measurement-system substitution; tolerance mapping; inspection decision; conformity assessment

## 1. Introduction

Coordinate inspection connects manufacturing variation to the decision to accept, rework, or reject a part. Fixed CMMs offer a controlled measuring volume, programmable probing, and stable fixturing. Portable articulated-arm CMMs can be taken to large or difficult-to-fixture components and used near the manufacturing process. Because the two systems serve overlapping inspection needs, shops may consider substituting one for the other. That substitution is safe only when it preserves the decision for the feature being inspected.

Instrument specifications alone do not answer this question. A maximum permissible error or verification result describes performance under stated test conditions, whereas a workpiece decision depends on the measured value, nominal value, tolerance, measurement procedure, and measurement uncertainty. A persistent difference between two systems can be irrelevant for a wide tolerance and decisive for a tight one. Interchangeability must therefore be judged at the feature and decision level rather than inferred from nominal accuracy.

ISO 10360-2 and ISO 10360-12 address acceptance and reverification of fixed and articulated coordinate measuring systems [1,3], while ISO 10360-5 covers contacting probing systems [2]. ISO 14253-1 separates instrument verification from conformity decisions on workpieces [4]. JCGM 100 and JCGM 106 provide the corresponding framework for uncertainty evaluation and decision risk [5,6]. These documents establish the principles required for defensible inspection, but they do not directly identify the tolerance range in which two recorded systems would assign different labels to the same feature.

Recent coordinate-metrology research has approached this problem from several directions. Sepahi-Boroujeni et al. [7] treated covariance in on-machine probing uncertainty, while Gąska et al. [8] proposed a task-specific assessment for articulated arms. Bisterov et al. [9] showed how measurement can be integrated with a machine tool, and Gao et al. [11] reviewed calibration, error modelling, uncertainty, and compensation. Taken together, these studies make one point clear: a measurement result cannot be separated from the architecture, calibration state, and procedure that produced it.

For portable systems, kinematic calibration, measurement posture, operator influence, and environmental conditions are especially relevant. Ibaraki and Saito [10] and Song et al. [15] addressed kinematic and posture-dependent calibration; Sun et al. [16] examined human-factor resilience; and Samel and Jankovych [17] studied ambient-temperature effects. Related work has assessed on-machine probing traceability [12], form-deviation conformity [13], and whether an articulated arm is capable for a defined task [14]. Recent CMM studies likewise emphasize task-specific uncertainty rather than instrument specifications alone [18,19].

Those studies provide the evidence needed to judge the quality or capability of an individual measurement system. They do not, however, directly answer a simpler shop-floor question that arises once two results are already in hand: at what tolerance would the choice of system change the direct inspection label? The mean difference between the instruments is not enough, because the answer also depends on the position of each result relative to nominal.

The measurements examined here came from a broader metrology and manufacturing-integrated digital-twin project [20,21]. The earlier work focused on data integration and prediction. Our question is narrower. We derive the tolerance interval in which the fixed-CMM and articulated-arm means receive opposite direct labels, relate the width of that interval to the raw inter-system difference, and extend the same logic to guard-banded decisions. The consolidated records do not support a claim of instrument bias or equivalence, so the analysis is confined to the decision consequences of the recorded values.

## 2. Materials and methods

### 2.1 Measurement data and profile selection

The analysis uses a consolidated workbook containing cylindrical, cubic, and spherical characteristics. For each characteristic, the file lists two fixed-CMM values and two FARO articulated-arm values under nominal condition labels of 20 °C and 30 °C. Project documentation identifies the portable system as a FARO Quantum S articulated arm and shows a Brown & Sharpe fixed CMM. It also reports 20 ± 0.5 °C and 30 ± 0.5 °C test conditions, humidity control, randomized measurement order, standardized fixturing, four-hour conditioning, and calibration checks [21]. These procedures cannot be tied independently to every workbook value because the native machine reports and environmental logs are not included. We therefore treat 20 °C and 30 °C as recorded condition labels and make no claim about a causal thermal effect or model-specific performance.

The dissertation describes a broader campaign with 20 parts and repeated measurements [21], but the workbook available for this paper does not preserve independent part-level variation. It contains 81 part-characteristic rows under 20 identifiers, yet the values repeat within each geometry-characteristic group. Collapsing those repeats produced 12 unique profiles. Two profiles concerned surface roughness and were excluded because the records do not identify the profilometer, stylus, cutoff, filter, or evaluation length. Six dimensional and four form profiles remained.

Other configuration details are also missing from the workbook, including probe geometry, datum alignment, point distribution, fitting and filtering settings, native report identifiers, independent repositioning, and task-specific uncertainty budgets. These choices can affect both dimensional and form results. The comparison is therefore between the two recorded result streams, not a verification of the accuracy, repeatability, or thermal response of either instrument model. Figure 1 shows the laboratory systems associated with the wider project; it does not establish that every stored value came from the exact configuration shown.

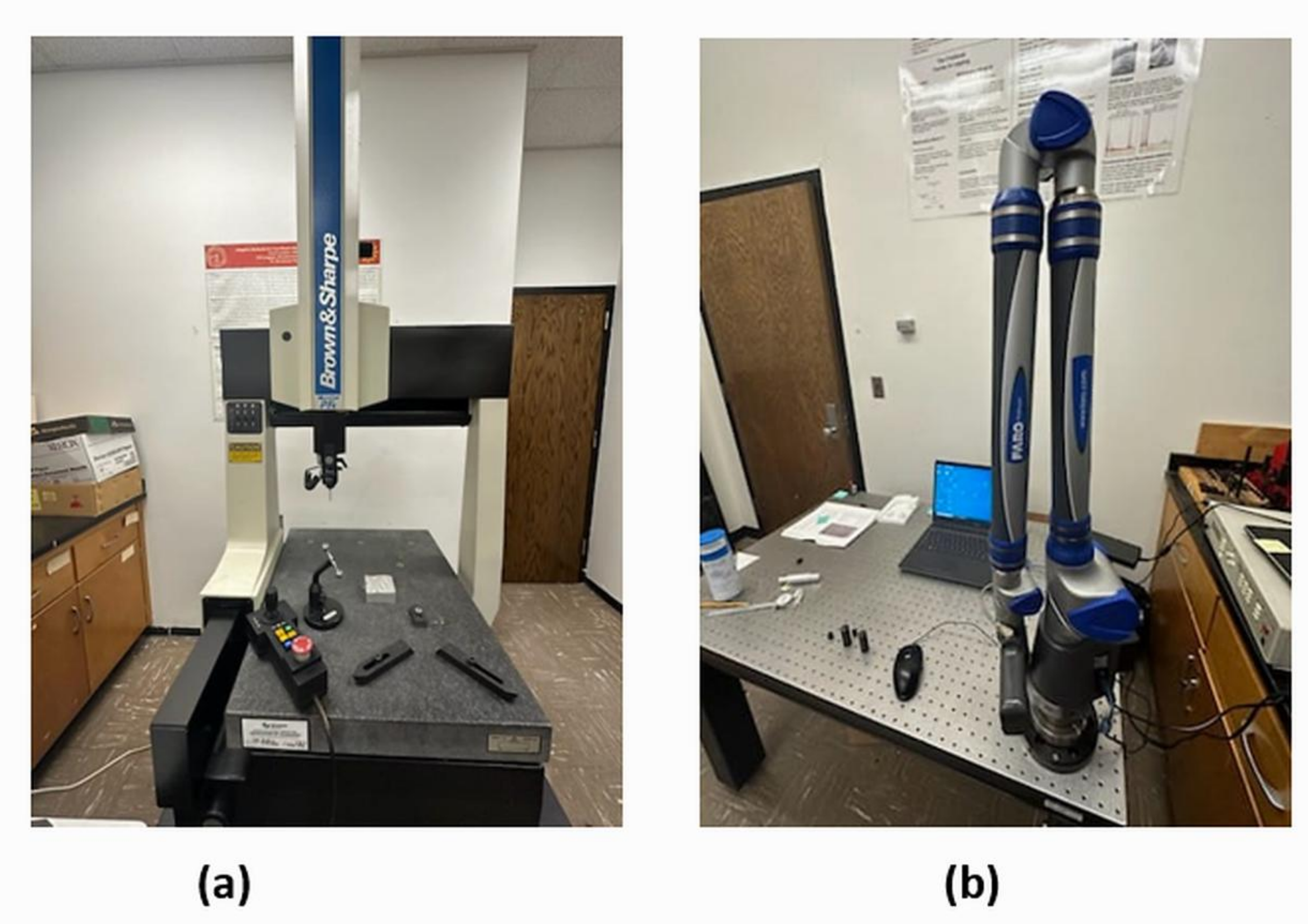

**Figure 1. Laboratory coordinate measurement systems associated with the broader project: (a) Brown & Sharpe fixed CMM and (b) FARO articulated-arm systems. The consolidated workbook does not retain native machine-report identifiers that link every stored value to a specific instrument configuration.**

## 2.2 Tolerance-based comparison

For system $s \in \{C, A\}$, profile j, condition label $c \in \{20, 30\}$, and stored value $r \in \{1, 2\}$, the mean of the two recorded values is

$$\bar{y}(s, j, c) = [y(s, j, c, 1) + y(s, j, c, 2)] / 2. \quad (1)$$

The value R = 2 describes the workbook structure and does not imply two independent experimental replications. The signed fixed-CMM minus articulated-arm difference, expressed in micrometres, is

$$d(j, c) = 1000[\bar{y}(C, j, c) - \bar{y}(A, j, c)]. \quad (2)$$

For each dimensional profile, both system means were referenced to the same nominal value N(j). Their absolute deviations from nominal are

$$e(s, j, c) = 1000\,|\bar{y}(s, j, c) - N(j)|. \quad (3)$$

For a bilateral tolerance with half-width $t \geq 0$, a direct arithmetic label was assigned according to

$$q(s, j, c; t) = 1 \text{ if } e(s, j, c) \leq t, \text{ and } 0 \text{ otherwise}. \quad (4)$$

Here q = 1 means that the recorded mean lies inside the nominal ±t window. This is not an ISO 14253-1 conformity decision because the available data do not contain a task-specific uncertainty budget or a guard band [4,6]. The indicator is used only to locate the tolerance values at which the two recorded means receive different arithmetic labels.

$$D(j, c) = [\min(eC, eA), \max(eC, eA)). \quad (5)$$

The lower endpoint is included because the first result enters the tolerance window at that value. The upper endpoint is excluded because both results are inside the window once the larger deviation is reached.

For the six dimensional profiles, the union Dany(c) contains tolerance values that affect at least one profile, while the intersection Dall(c) contains tolerance values that affect all six. The proportion of profiles with opposite labels is

$$P(c, t) = (1/6)\,\Sigma\,\delta(j, c; t), \quad j \in J_D. \quad (6)$$

The width of the profile-specific disagreement interval is

$$WD(j, c) = |eC - eA| \leq 1000\,|\bar{y}C - \bar{y}A|. \quad (7)$$

The inequality follows from the reverse triangle inequality. The two quantities are equal when both results lie on the same side of nominal. When the results straddle nominal, the raw inter-system difference is larger than the tolerance range over which their labels disagree.

If defensible system-specific guard bands G(s,j,c) become available, the same formulation can be applied after replacing e(s,j,c) with e(s,j,c) + G(s,j,c). This retains the interval structure while respecting the relevant uncertainty-based decision rule. No numerical guard-band result is reported here because the necessary uncertainty budgets are unavailable.

For cylindricity, flatness, perpendicularity, and sphericity, the workbook records a one-sided upper limit U = 10 µm. Their direct threshold labels were obtained from $q_F(s,j,c) = 1$ when $f(s,j,c) \leq U$ and $q_F(s,j,c) = 0$ otherwise. Changes between the two condition labels were described arithmetically and were not interpreted as thermal coefficients.

Figure 2 summarizes the five-step analytical procedure used in this study. The two recorded measurement streams were first organized by geometry, characteristic, and nominal condition label. Repeated values within each geometry-characteristic group were then consolidated, and two surface-roughness profiles were excluded because the available records did not contain sufficient acquisition information. For each retained profile, the recorded values were averaged separately for the fixed CMM and articulated-arm CMM. Their deviations from nominal were subsequently used to identify the tolerance range over which the two recorded means receive different direct inspection labels. The final step summarizes the dimensional and form results obtained from the retained profiles.

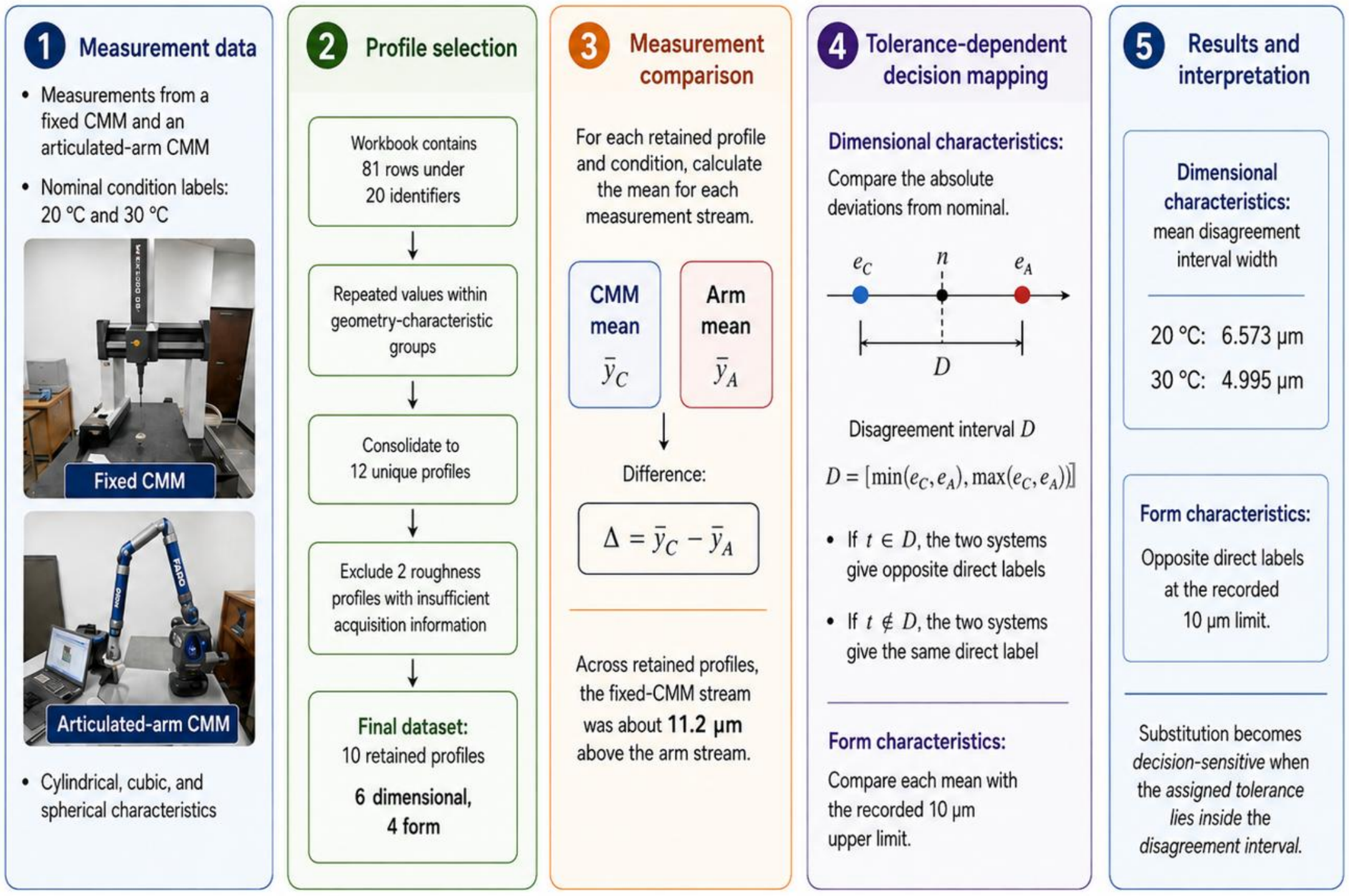

**Figure 2. Five-step workflow for the tolerance-dependent comparison of the fixed-CMM and articulated-arm CMM measurement streams. Repeated workbook entries were consolidated into unique profiles, the retained system means were compared, and their deviations from nominal were mapped to the tolerance conditions under which the recorded means receive different direct inspection labels.**

## 3. Results

### 3.1 Differences between the fixed CMM and articulated arm

The fixed-CMM mean was higher for every retained profile at both condition labels. Across profiles, the mean difference was 11.175 μm at 20 °C and 11.202 μm at 30 °C, with standard deviations of 0.035 and 0.039 μm, respectively. The positive sign is consistent throughout Figure 3, and the variation among profiles is small compared with the overall separation.

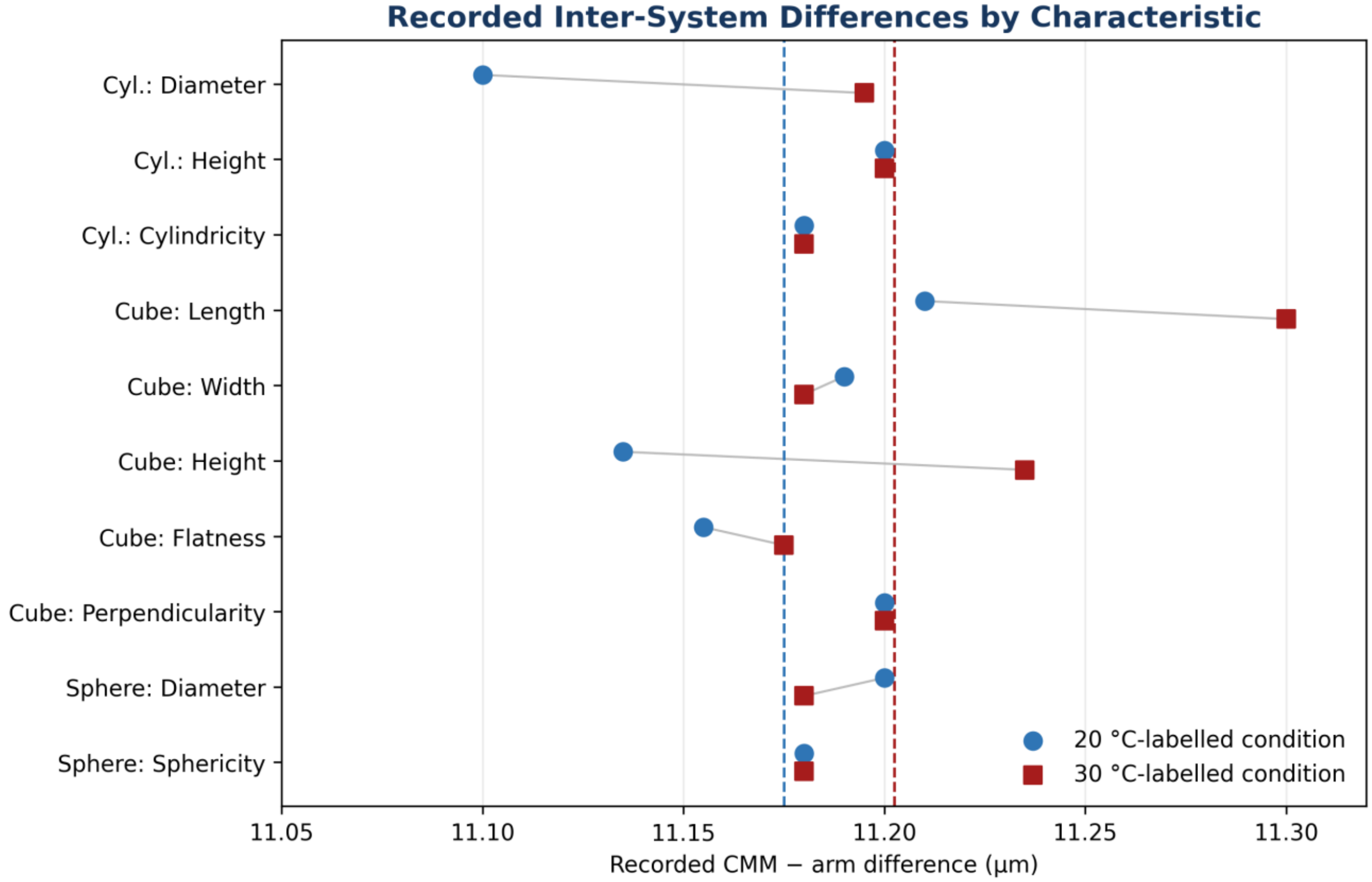

**Figure 3. Recorded fixed-CMM minus articulated-arm differences for the ten retained characteristics. Dashed lines mark the across-profile mean difference at each condition label. The values are paired recorded differences, not traceable bias estimates.**

The paired entries within each system-profile-condition cell differ very little: the mean absolute separation is 0.029 µm for the fixed CMM and 0.031 µm for the articulated arm. Because the workbook does not document independent setup, repositioning, or operator conditions, these values are not estimates of instrument repeatability. Between the 20 °C and 30 °C labels, the fixed-CMM means rise by 0.800 µm on average and the arm means by 0.772 µm. Their similar changes leave the inter-system gap close to 11.2 µm.

The same pattern appears for all four form characteristics. At both conditions, the fixed-CMM mean lies above the recorded 10 µm upper limit and the articulated-arm mean lies below it, giving opposite direct labels in all eight characteristic-condition pairs. Figure 4 presents these threshold positions. It does not indicate which instrument is correct and does not apply an uncertainty-aware conformity rule.

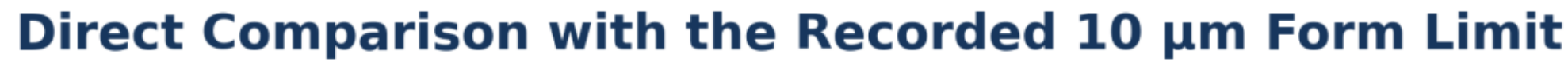

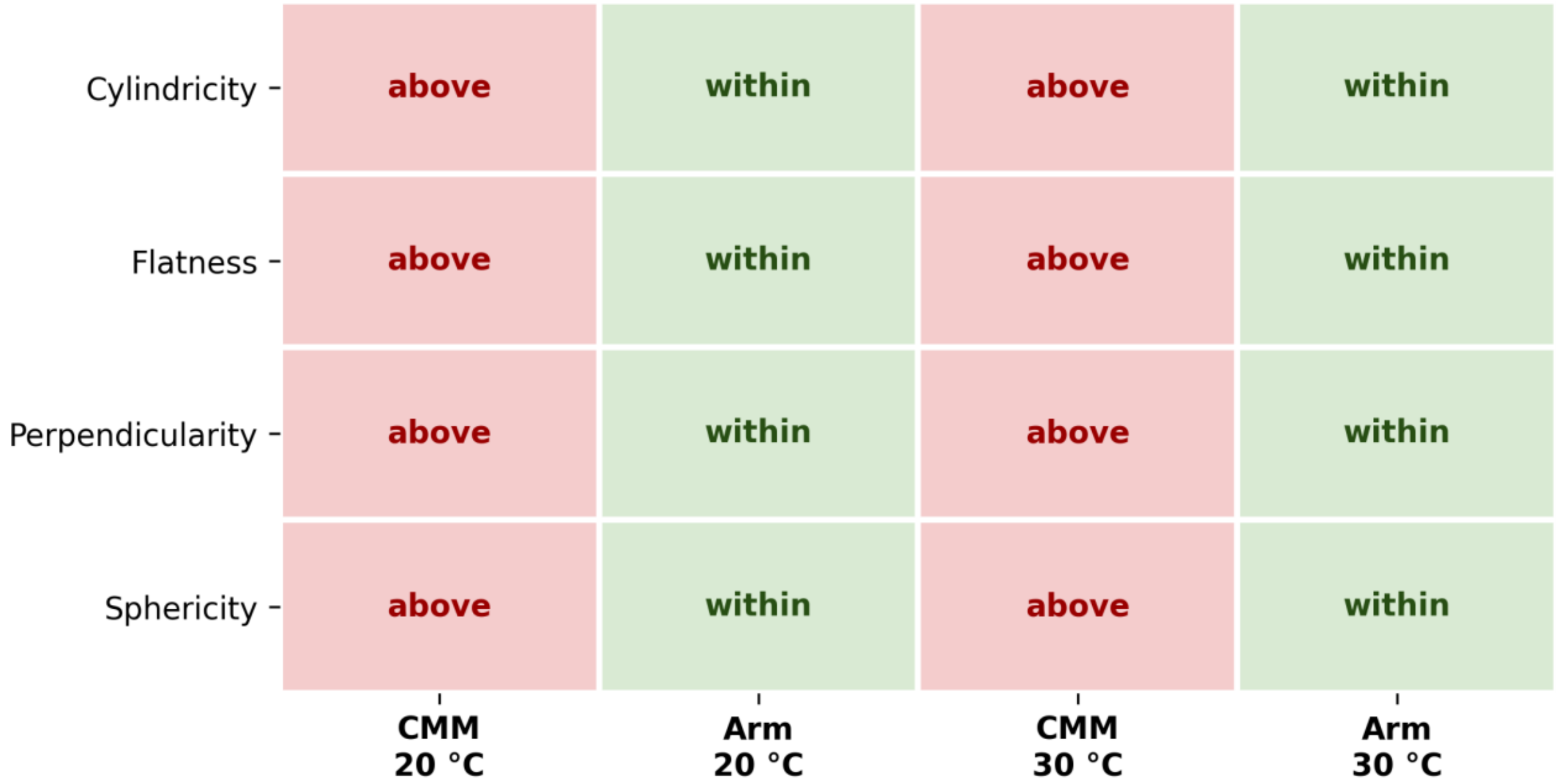

Arithmetic threshold positions only; not uncertainty-aware conformity decisions.

**Figure 4. Direct positions of the retained form measurements relative to the recorded 10 µm upper limit. The cells show arithmetic threshold positions only; no uncertainty-based conformity rule or guard band is applied.**

### 3.2 Tolerance-dependent disagreement

Each dimensional characteristic yields a finite disagreement interval. At 20 °C, the lower endpoints lie between 2.275 and 2.335 µm and the upper endpoints between 8.825 and 8.900 µm. At 30 °C, the corresponding ranges are 3.080–3.165 µm and 8.090–8.135 µm. Figure 5 shows the strong overlap among profiles at each condition.

Exact Dimensional Threshold-Disagreement Intervals

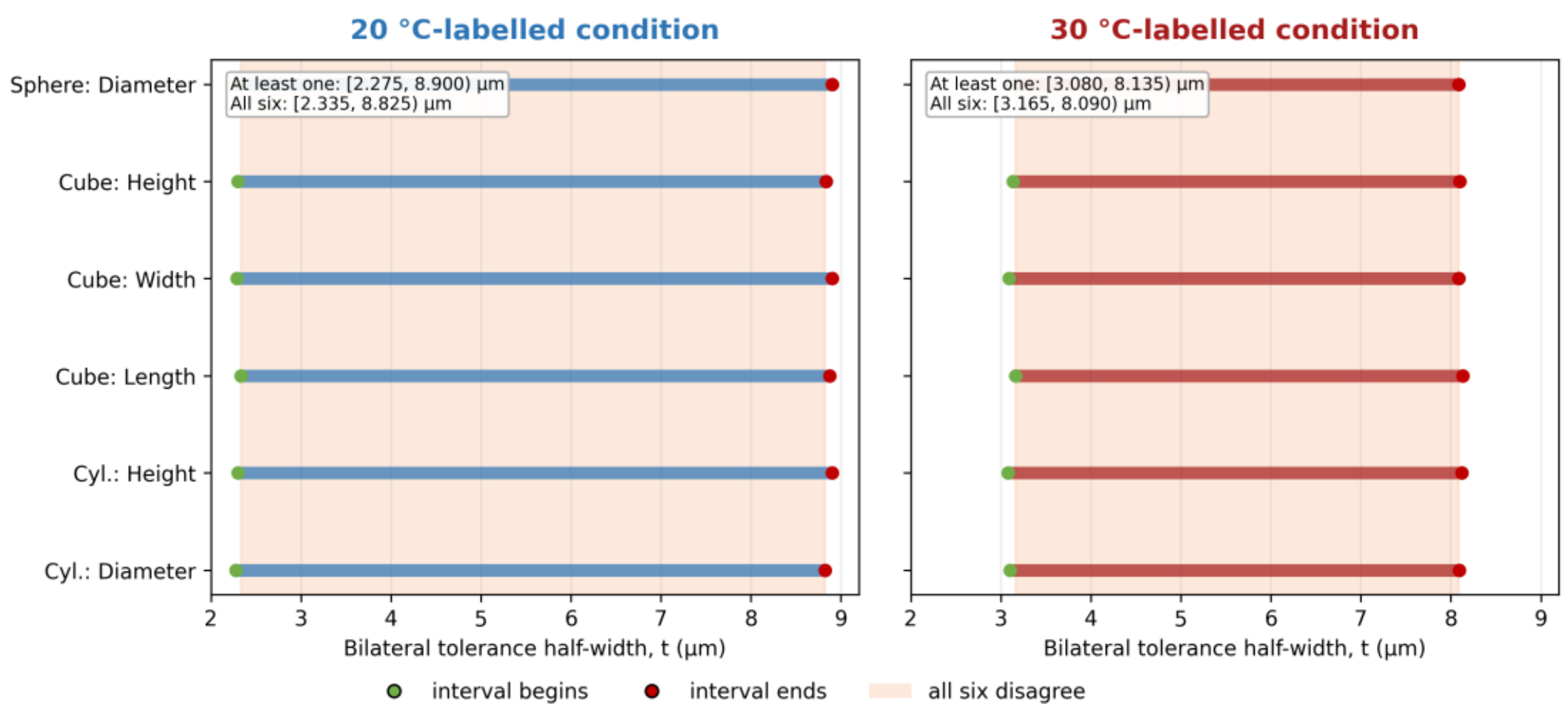

**Figure 5. Bilateral tolerance half-width intervals that produce opposite direct labels for the six dimensional characteristics. The shaded region marks the shared interval in which all six characteristics disagree.**

**Table 1. Profile-specific bilateral disagreement intervals and decision-disagreement widths.**

| Geometry | Characteristic | 20 °C interval (µm) | Width at 20 °C (µm) | 30 °C interval (µm) | Width at 30 °C (µm) |
|---|---|---|---|---|---|
| Cylinder | Diameter | [2.275, 8.825) | 6.550 | [3.100, 8.095) | 4.995 |
| Cylinder | Height | [2.300, 8.900) | 6.600 | [3.080, 8.120) | 5.040 |
| Cube | Length | [2.335, 8.875) | 6.540 | [3.165, 8.135) | 4.970 |
| Cube | Width | [2.290, 8.900) | 6.610 | [3.090, 8.090) | 5.000 |
| Cube | Height | [2.300, 8.835) | 6.535 | [3.135, 8.100) | 4.965 |
| Sphere | Diameter | [2.300, 8.900) | 6.600 | [3.090, 8.090) | 5.000 |

The mean interval width is 6.573 µm at 20 °C and 4.995 µm at 30 °C. The decision-sensitive range therefore becomes narrower at the second condition label, although the average raw difference between systems changes very little.

The overlapping profile intervals also define useful aggregate regions. At 20 °C, at least one characteristic disagrees for $2.275 \le t < 8.900$ µm, and all six disagree for $2.335 \le t < 8.825$ µm. At 30 °C, the corresponding regions are $3.080 \le t < 8.135$ µm and $3.165 \le t < 8.090$ µm. Table 2 lists these ranges; Figure 6 gives the proportion of dimensional characteristics with opposite labels across t.

**Table 2. Aggregate direct-decision regions for the retained dimensional characteristics.**

| Condition label | Same label below the union | At least one characteristic disagrees | All six characteristics disagree |
|---|---|---|---|
| 20 °C | $0 \le t < 2.275$ µm | $2.275 \le t < 8.900$ µm | $2.335 \le t < 8.825$ µm |
| 30 °C | $0 \le t < 3.080$ µm | $3.080 \le t < 8.135$ µm | $3.165 \le t < 8.090$ µm |

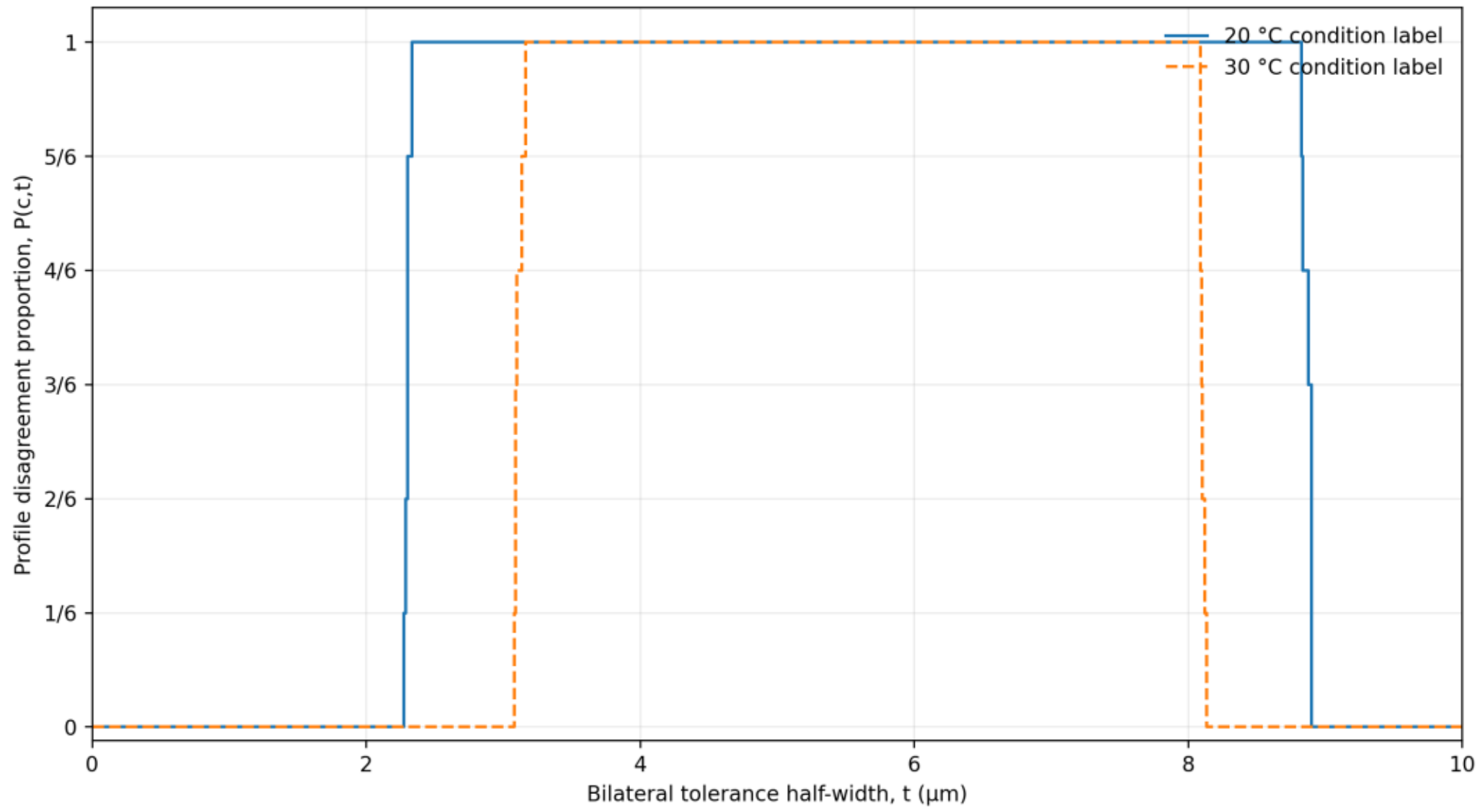

**Figure 6. Proportion P(c,t) of dimensional characteristics with opposite direct labels. A plateau at one marks the tolerance range in which all six retained characteristics change label between the two systems.**

For tolerance half-widths below the smallest lower endpoint, both recorded means lie outside the window. At or above the largest upper endpoint, both lie inside. The resulting agreement is limited to the arithmetic label for those recorded values; it is not evidence of measurement equivalence.

## 4. Discussion

The approximately 11.2 μm separation is only the starting point. Its inspection consequence depends on where the two readings sit relative to nominal. Equation (5) captures that consequence directly: the interval between the smaller and larger absolute deviations is the only range of tolerance half-widths for which the labels differ. The width of this interval may equal the raw system difference, but only when both readings are on the same side of nominal. If they straddle nominal, part of the separation lies across the nominal value and does not enlarge the disagreement range. This explains why the raw difference remained nearly unchanged across the two condition labels while the mean disagreement width fell from 6.573 to 4.995 μm.

The form characteristics provide the clearest practical example. For cylindricity, flatness, perpendicularity, and sphericity, the fixed-CMM and articulated-arm means fell on opposite sides of the recorded 10 μm limit at both conditions. A shop that alternated between the systems would therefore obtain different direct labels for every retained form profile. The records do not reveal which result is closer to the measurand, because no traceable reference and no fully matched measurement procedure are available.

This tolerance view sits alongside, rather than in place of, established uncertainty and capability methods. Full-covariance and task-specific uncertainty analyses support defensible conformity decisions [7,18]. Calibration and error-compensation studies improve the quality of the result stream [10,11], and capability studies determine whether a system is suitable for a defined task [8,14]. The interval developed here answers an earlier planning question: where is the observed difference large enough, relative to the assigned tolerance, to justify that deeper validation?

For a manufacturing engineer, the calculation is straightforward. The actual drawing tolerance is placed against the interval for the characteristic. A tolerance inside the interval marks a decision-sensitive task, so substitution should wait until the measurand, probing strategy, software settings, and uncertainty budgets have been matched and an allowable difference has been specified. A tolerance outside the interval means that the two recorded means retain the same arithmetic label. It should not be mistaken for proof of equivalence.

The treatment of the workbook is equally important to the result. Although the file lists 20 part identifiers, the values repeat within each geometry-characteristic group. Counting those rows as independent observations would create pseudoreplication. We therefore used the unique profile as the analytical unit. This choice also sets the limits of the paper: the workbook does not preserve native machine reports, probe geometry, datum alignment, point patterns, fitting and filtering settings, independent setup repetitions, traceable reference values, or numerical uncertainty budgets. The findings describe a small deterministic record and cannot be generalized across instruments, operators, parts, or facilities.

A controlled comparison can test the method more rigorously. Such a study should use calibrated artifacts or independently traceable features, common measurand definitions and software settings, randomized measurement order, independent repositioning, and a pre-specified equivalence margin. Artifact and environmental temperatures should be logged, and replication should be selected from the required precision or statistical power. The resulting analysis should report task-specific uncertainty, agreement at the actual drawing tolerances, and producer and consumer risk where relevant. That experiment would show whether the intervals derived from paired means predict the decision changes observed in practice.

## 5. Conclusions

For a dimensional characteristic, the distances of two system means from nominal define a half-open interval of tolerance values for which the direct labels disagree. The interval width is no greater than the raw inter-system difference and equals that difference only when both readings lie on the same side of nominal. This converts a numerical separation into a tolerance-specific statement about decision stability.

In the retained data, the fixed-CMM values were about 11.2 μm higher than the articulated-arm values at both condition labels. All four form characteristics fell on opposite sides of the recorded 10 μm limit, while the six dimensional characteristics produced strongly overlapping disagreement intervals with mean widths of 6.573 μm at 20 °C and 4.995 μm at 30 °C. These results do not identify the correct instrument or certify interchangeability. They show where instrument choice can alter the recorded inspection label and where a focused equivalence study and task-specific uncertainty budget should precede substitution.

## Data availability

The measurement workbook is not publicly deposited. Requests for access may be directed to the corresponding author.

## Declaration of competing interest

The authors declare that they have no known competing financial interests or personal relationships that could have appeared to influence the work reported in this paper.

### Declaration of generative AI and AI-assisted technologies in the manuscript preparation process

During preparation of this work, the authors used AI to assist with language editing, organization, and consistency checking. The authors reviewed and edited the manuscript and take full responsibility for its content.